\documentclass[10pt,twocolumn,letterpaper]{article}

\usepackage{cvpr}
\usepackage{times}
\usepackage{epsfig}
\usepackage{graphicx}
\usepackage{amsmath}
\usepackage{amssymb}

\usepackage[utf8]{inputenc} 
\usepackage[T1]{fontenc}    
\usepackage{url}            
\usepackage{booktabs}       
\usepackage{amsfonts}       
\usepackage{nicefrac}       
\usepackage{microtype}      

\usepackage{color}
\usepackage{adjustbox}
\usepackage{multirow}
\usepackage{subcaption}

\usepackage{bbold}

\newcommand{\beq}{\vspace{0mm}\begin{equation}}
\newcommand{\eeq}{\vspace{0mm}\end{equation}}
\newcommand{\beqs}{\vspace{0mm}\begin{eqnarray}}
\newcommand{\eeqs}{\vspace{0mm}\end{eqnarray}}
\newcommand{\barr}{\begin{array}}
\newcommand{\earr}{\end{array}}

\newcommand{\Cmat}{{\bf C}}

\newcommand{\Imat}{{\bf I}}

\newcommand{\Umat}[0]{{{\bf U}}}
\newcommand{\Vmat}[0]{{{\bf V}}}
\newcommand{\Wmat}[0]{{{\bf W}}}
\newcommand{\Xmat}[0]{{{\bf X}}}

\newcommand{\cv}[0]{{\boldsymbol{c}}}

\newcommand{\fv}[0]{{\boldsymbol{f}}}

\newcommand{\hv}[0]{{\boldsymbol{h}}}
\newcommand{\iv}[0]{{\boldsymbol{i}}}

\newcommand{\ov}[0]{{\boldsymbol{o}}}

\newcommand{\sv}[0]{{\boldsymbol{s}}}

\newcommand{\vv}{\boldsymbol{v}}
\newcommand{\wv}{\boldsymbol{w}}
\newcommand{\xv}{\boldsymbol{x}}
\newcommand{\yv}{\boldsymbol{y}}
\newcommand{\zv}{\boldsymbol{z}}

\newcommand{\R}{\mathbb{R}}

\newcommand{\Tcal}{\mathcal{T}}

\newcommand{\Hcal}{\mathcal{H}}

\usepackage[pagebackref=true,breaklinks=true,letterpaper=true,colorlinks,bookmarks=false]{hyperref}

\cvprfinalcopy 

\def\cvprPaperID{2390} 

\ifcvprfinal\fi
\begin{document}

\title{Semantic Compositional Networks for Visual Captioning}

\author{Zhe Gan$^\dag$, Chuang Gan$^*$, Xiaodong He$^\ddag$, Yunchen Pu$^\dag$ \\
	Kenneth Tran$^\ddag$, Jianfeng Gao$^\ddag$, Lawrence Carin$^\dag$, Li Deng$^\ddag$\\
$^\dag$Duke University, $^*$Tsinghua University, $^\ddag$Microsoft Research, Redmond, WA 98052, USA\\
{\tt\small \{zhe.gan, yunchen.pu, lcarin\}@duke.edu},
{\tt\small ganchuang1990@gmail.com} \\
{\tt\small \{xiaohe, ktran, jfgao, deng\}@microsoft.com}
}

\maketitle

\begin{abstract}
A Semantic Compositional Network (SCN) is developed for image captioning, in which semantic concepts ($i.e.$, tags) are detected from the image, and the probability of each tag is used to compose the parameters in a long short-term memory (LSTM) network. The SCN extends each weight matrix of the LSTM to an ensemble of tag-dependent weight matrices. The degree to which each member of the ensemble is used to generate an image caption is tied to the image-dependent probability of the corresponding tag. In addition to captioning images, we also extend the SCN to generate captions for video clips. We qualitatively analyze semantic composition in SCNs, and quantitatively evaluate the algorithm on three benchmark datasets: COCO, Flickr30k, and Youtube2Text. Experimental results show that the proposed method significantly outperforms prior state-of-the-art approaches, across multiple evaluation metrics.
\end{abstract}

\section{Introduction}
There has been a recent surge of interest in developing models that can generate captions for images or videos, termed visual captioning. Most of these approaches learn a probabilistic model of the caption, conditioned on an image or a video~\cite{mao2014deep,venugopalan2014translating,fang2015captions,karpathy2015deep,vinyals2015show,xu2015show,donahue2015long,venugopalan2015sequence,pan2016joint,yu2016video}. Inspired by the successful use of the encoder-decoder framework employed in machine translation~\cite{bahdanau2014neural,cho2014learning,sutskever2014sequence}, most recent work on visual captioning employs a convolutional neural network (CNN) as an encoder, obtaining a fixed-length vector representation of a given image or video. A recurrent neural network (RNN), typically implemented with long short-term memory (LSTM) units~\cite{hochreiter1997long}, is then employed as a decoder to generate a caption. 

\begin{figure}[t]
	\centering
	\begin{subfigure}{.48\textwidth}
		\centering
		\includegraphics[width=1.0\linewidth]{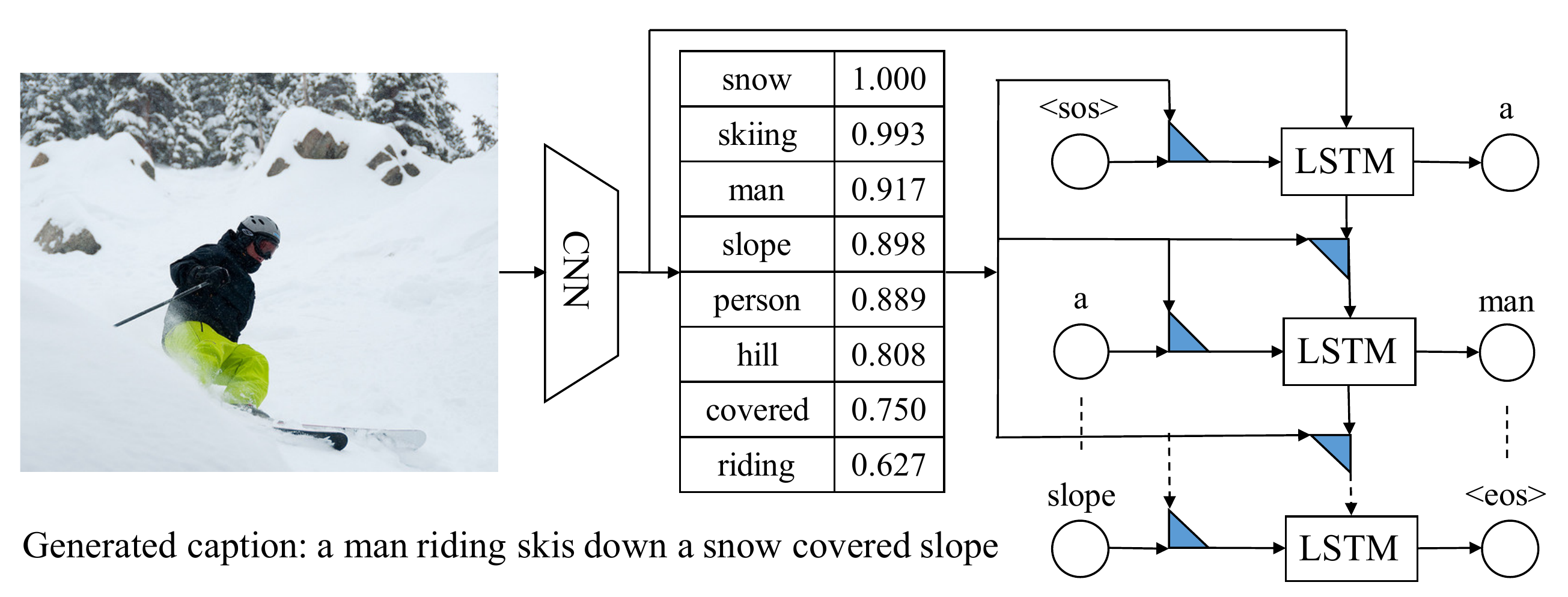}
		\caption{\small{Overview of the proposed model.}}
		\label{fig:overview}
	\end{subfigure}
	\begin{subfigure}{.48\textwidth}
		\centering
		\includegraphics[width=0.95\linewidth]{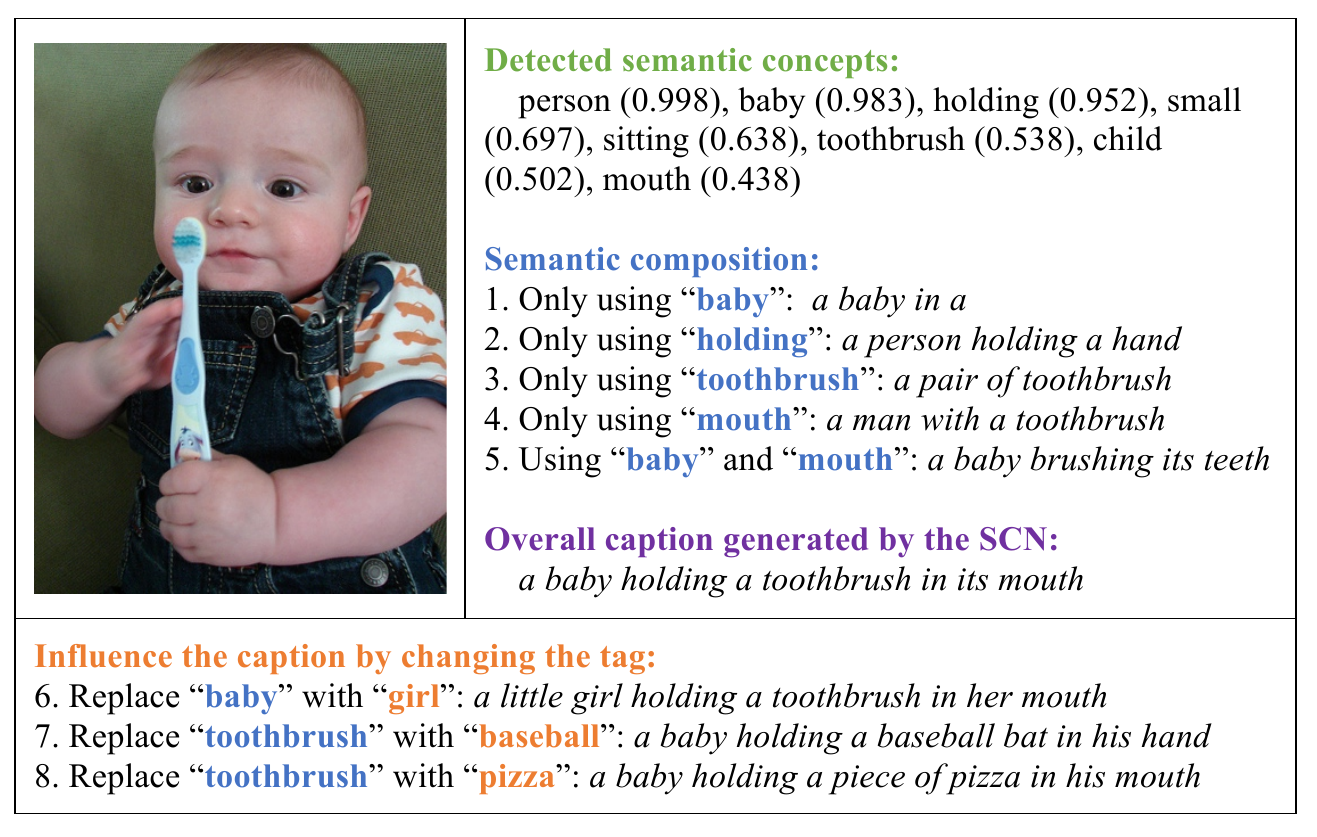}
		\caption{\small{Examples of SCN-based image captioning.}}
		\label{fig:illustration}
	\end{subfigure}
	\caption{\small{Model architecture and illustration of semantic composition. Each triangle symbol represents an ensemble of tag-dependent weight matrices. The number next to a semantic concept (\emph{i.e.}, a tag) is the probability that the corresponding semantic concept is presented in the input image.}}
	\label{fig:framework}
	\vspace{-4mm}
\end{figure}

Recent work shows that adding explicit high-level semantic concepts (\emph{i.e.}, tags) of the input image/video can further improve visual captioning. 
As shown in~\cite{wu2015value,you2016image}, detecting explicit semantic concepts encoded in an image, and adding this high-level semantic information into the CNN-LSTM framework, has improved performance significantly. 
Specifically,~\cite{wu2015value} feeds the semantic concepts as an \emph{initialization} step into the LSTM decoder. In~\cite{you2016image}, a model of semantic attention is proposed which selectively attends to semantic concepts through a soft attention mechanism~\cite{bahdanau2014neural}. 
On the other hand, although significant performance improvements were achieved, integration of semantic concepts into the LSTM-based caption generation process is constrained in these methods; \emph{e.g.}, only through soft attention or initialization of the first step of the LSTM.


In this paper, we propose the Semantic Compositional Network (SCN) to more effectively assemble the meanings of individual tags to generate the caption that describes the overall meaning of the image, as illustrated in Figure~\ref{fig:overview}.
Similar to the conventional CNN-LSTM-based image captioning framework, a CNN is used to extract the visual feature vector, which is then fed into a LSTM for generating the image caption (for simplicity, in this discussion we refer to images, but the method is also applicable to video). 
However, unlike the conventional LSTM, the SCN extends each weight matrix of the conventional LSTM to an {\em ensemble} of tag-dependent weight matrices, subject to the probabilities that the tags are present in the image. These tag-dependent weight matrices form a weight tensor with a large number of parameters. In order to make learning feasible, we factorize that tensor to be a three-way matrix product, which dramatically reduces the number of free parameters to be learned, while also yielding excellent performance. 


The main contributions of this paper are as follows: (\emph{i}) We propose the SCN to effectively compose individual semantic concepts for image captioning. (\emph{ii}) We perform comprehensive evaluations on two image captioning benchmarks, demonstrating that the proposed method outperforms previous state-of-the-art approaches by a substantial margin. For example, as reported by the COCO official test server, we achieve a BLEU-4 of 33.1, an improvement of 1.5 points over the current published state-of-the-art ~\cite{you2016image}. (\emph{iii}) We extend the proposed framework from image captioning to video captioning, demonstrating the versatility of the proposed model. (\emph{iv}) We also perform a detailed analysis to study the SCN, showing that the model can adjust the caption smoothly by modifying the tags.

\section{Related work}
%

%
We focus on recent neural-network-based literature for caption generation, as these are most relevant to our work. Such models typically extract a visual feature vector via a CNN, and then send that vector to a language model for caption generation. Representative works include~\cite{chen2015mind,devlin2015language,donahue2015long,karpathy2015deep,kiros2014multimodal,kiros2014unifying,mao2014deep,vinyals2015show} for image captioning and ~\cite{donahue2015long,venugopalan2015sequence,venugopalan2014translating,yu2016video,ballas2015delving,pu2016adaptive,dong2016early} for video captioning. The differences of the various methods mainly lie in the types of CNN architectures and language models. For example, the vanilla RNN~\cite{elman1990finding} was used in~\cite{mao2014deep,karpathy2015deep}, while the LSTM~\cite{hochreiter1997long} was used in~\cite{vinyals2015show,venugopalan2015sequence,venugopalan2014translating}. The visual feature vector was only fed into the RNN once at the first time step in~\cite{vinyals2015show,karpathy2015deep}, while it was used at each time step of the RNN in~\cite{mao2014deep}.

Most recently, \cite{xu2015show} utilized an attention-based mechanism to learn where to focus in the image during caption generation. This work was followed by~\cite{yang2016encode} which introduced a review module to improve the attention mechanism and~\cite{liu2016attention} which proposed a method to improve the correctness of visual attention. Moreover, a variational autoencoder was developed in~\cite{VAE_NIPS2016} for image captioning. Other related work includes~\cite{pan2016joint} for video captioning and~\cite{anne2016deep} for composing sentences that describe novel objects.


Another class of models uses semantic information for caption generation. Specifically, ~\cite{jia2015guiding} applied retrieved sentences as additional semantic information to guide the LSTM when generating captions, while~\cite{fang2015captions,wu2015value,you2016image} applied a semantic-concept-detection process~\cite{gan2016learning} before generating sentences. In addition,~\cite{fang2015captions} also proposes a deep multimodal similarity model to project visual features and captions into a joint embedding space.
This line of methods represents the current state of the art for image captioning. 
Our proposed model also lies in this category; however, distinct from the aforementioned approaches, 
our model uses weight \emph{tensors} in LSTM units. This allows learning an ensemble of semantic-concept-dependent weight matrices for generating the caption.
%
%

Related to but distinct from the hierarchical composition in a recursive neural network~\cite{socher2014grounded}, our model carries out implicit composition of concepts, and there is no hierarchical relationship among these concepts.
Figure~\ref{fig:illustration} illustrates the semantic composition manifested in the SCN model. Specifically, a set of semantic concepts, such as ``\emph{baby, holding, toothbrush, mouth}'', are detected with high probabilities. If only one semantic concept is turned on, the model will generate a description covering only part of the input image, as shown in sentences 1-5 of Figure~\ref{fig:illustration}; however, by assembling all these semantic concepts, the SCN is able to generate a comprehensive description ``\emph{a baby holding a toothbrush in its mouth}''. More interestingly, as shown in sentences 6-8 of Figure~\ref{fig:illustration}, the SCN also has great flexibility to adjust the generation of the caption by changing certain semantic concepts. 

The tensor factorization method is used to make the SCN compact and simplify learning.
Similar ideas have been exploited 
in~\cite{kiros2014multiplicative,memisevic2007unsupervised,song2016factored,sutskever2011generating,taylor2009factored,wu2016multiplicative,gan2017style}. 
%
In~\cite{donahue2015long,jin2015aligning,kiros2014multimodal} the authors also briefly discussed using the tensor factorization method for image captioning. Specifically, visual features extracted from CNNs are utilized in~\cite{donahue2015long,kiros2014multimodal}, and an inferred scene vector is used in~\cite{jin2015aligning} for tensor factorization. In contrast to these works, we use the semantic-concept vector that is formed by the probabilities of all tags to weight the basis LSTM weight matrices in the ensemble. 
Our semantic-concept vector is more powerful than the visual-feature vector~\cite{donahue2015long,kiros2014multimodal} and the scene vector~\cite{jin2015aligning} in terms of providing explicit semantic information of an image, hence leading to significantly better performance, as shown in our quantitative evaluation. In addition, the usage of semantic concepts also makes the proposed SCN more interpretable than~\cite{donahue2015long,jin2015aligning,kiros2014multimodal}, as shown in our qualitative analysis, since each unit in the semantic-concept vector corresponds to an explicit tag.

\section{Semantic compositional networks}


\subsection{Review of RNN for image captioning}\label{sec:img_cap_lstm}
Consider an image $\Imat$, with associated caption $\Xmat$. We first extract feature vector $\vv(\Imat)$, which is often the top-layer features of a pretrained CNN. Henceforth, for simplicity, we omit the explicit dependence on $\Imat$, and represent the visual feature vector as $\vv$.
The length-$T$ caption is represented as $\Xmat=(\xv_1, \ldots, \xv_{T})$, with $\xv_t$ a 1-of-$V$ (``one hot'') encoding vector, with $V$ the size of the vocabulary. The length $T$ typically varies among different captions. 

The $t$-th word in a caption, $\xv_t$, is linearly embedded into an $n_x$-dimensional real-valued vector $\wv_t=\Wmat_e \xv_t$, where $\Wmat_e \in \R^{n_x\times V}$ is a word embedding matrix (learned), \emph{i.e.}, $\wv_t$ is a column of $\Wmat_e$ chosen by the one-hot $\xv_t$. The probability of caption $\Xmat$ given image feature vector $\vv$ is defined as
\begin{align}
p(\Xmat|\Imat) = \textstyle{\prod_{t=1}^T} p(\xv_t|\xv_0,\ldots,\xv_{t-1},\vv) \,,
\end{align} 
where $\xv_0$ is defined as a special start-of-the-sentence token. All the words in the caption are sequentially generated using a RNN, until the end-of-the-sentence symbol is generated. Specifically, each conditional  $p(\xv_t|\xv_{<t},\vv)$ is specified as $\mbox{softmax}(\Vmat \hv_t)$, where $\hv_t$ is recursively updated through  $\hv_t =\Hcal(\wv_{t-1},\hv_{t-1}, \vv)$, and $\hv_{0}$ is defined as a zero vector ($\hv_0$ is {\em not} updated during training). $\Vmat$ is the weight matrix connecting the RNN's hidden state, used for computing a distribution over words. Bias terms are omitted for simplicity throughout the paper.

Without loss of generality, we begin by discussing an RNN with a simple transition function $\Hcal(\cdot)$; this is generalized in Section~\ref{sec:img_cap_f_lstm} to the LSTM. Specifically, $\Hcal(\cdot)$ is defined as 
\begin{align} \label{eq:basic_rnn}
\hv_t = \sigma (\Wmat \xv_{t-1} + \Umat \hv_{t-1} +\mathbb{1}(t=1) \cdot \Cmat \vv  ) \,,
\end{align}
where $\sigma(\cdot)$ is a logistic sigmoid function, and $\mathbb{1}(\cdot)$ represents an indicator function. Feature vector $\vv$ is fed into the RNN at the beginning, $i.e.$, at $t=1$. $\Wmat$ is defined as the input matrix, and $\Umat$ is termed the recurrent matrix. 
%
The model in (\ref{eq:basic_rnn}) is illustrated in Figure~\ref{fig:model}(a). 


\subsection{Semantic concept detection}\label{sec:visual_concept_detection}
The SCN developed below is based on the detection of semantic concepts, \emph{i.e.}, tags, in the image under test. In order to detect such from an image, we first select a set of tags from the caption text in the training set. Following~\cite{fang2015captions}, we use the $K$ most common words in the training captions to determine the vocabulary of tags, which includes the most frequent nouns, verbs, or adjectives. 

In order to predict semantic concepts given a test image, motivated by~\cite{wu2015value,tran2016rich}, we treat this problem as a multi-label classification task. Suppose there are $N$ training examples, and $\yv_i=[y_{i1},\ldots,y_{iK}]\in\{0,1\}^K$ is the label vector of the $i$-th image, where $y_{ik}=1$ if the image is annotated with tag $k$, and $y_{ik}=0$ otherwise. Let $\vv_i$ and $\sv_i$ represent the image feature vector and the semantic feature vector for the $i$-th image,
the cost function to be minimized is
\begin{align}
\frac{1}{N} \sum_{i=1}^N \sum_{k=1}^K \Big( y_{ik}\log s_{ik} + (1-y_{ik}) \log (1-s_{ik})  \Big) \,,
\end{align} 
where $\sv_i = \sigma\big(f(\vv_i)\big)$ is a $K$-dimensional vector with $\sv_i=[s_{i1},\ldots,s_{iK}]$, \,$\sigma(\cdot)$ is the logistic sigmoid function and $f(\cdot)$ is implemented as a multilayer perceptron (MLP).  

In testing, for each input image, we compute a semantic-concept vector $\sv$, formed by the probabilities of all tags, computed by the semantic-concept detection model. 

\subsection{SCN-RNN} \label{sec:img_cap_tensor_LSTM}
\begin{figure}[t]
	\centering
	\includegraphics[width=0.40\textwidth]{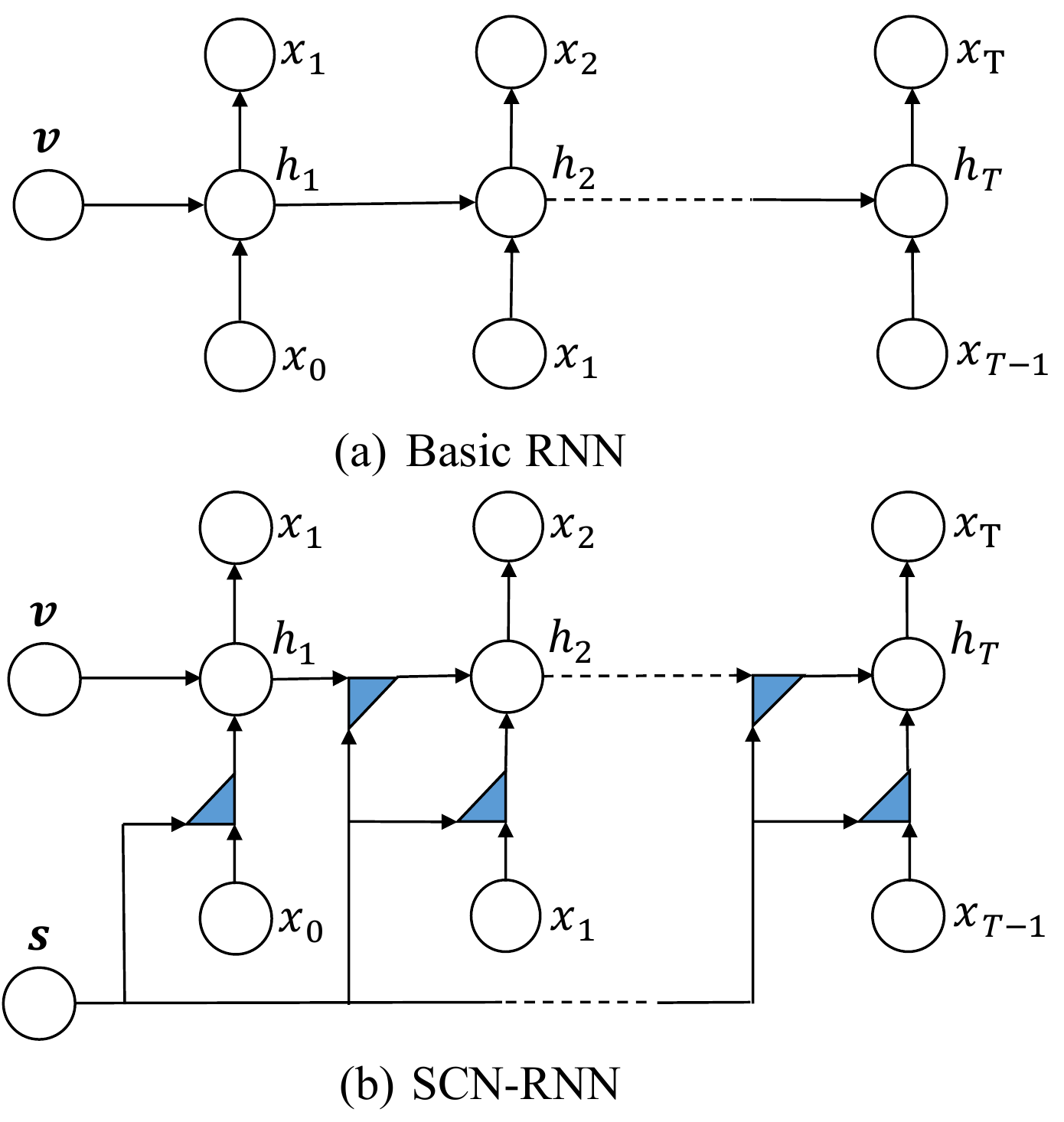}
	\vspace{-3mm}
	\caption{\small{Comparison of our proposed model with a conventional recurrent neural network (RNN) for caption generation. $\vv$ and $\sv$ denote the visual feature and semantic feature, respectively. $\xv_0$ represents a special start-of-the-sentence token, $(\xv_1,\ldots,\xv_T)$ represents the caption, and $(\hv_1,\ldots,\hv_T)$ denotes the RNN hidden states. Each triangle symbol represents an ensemble of tag-dependent weight matrices.}}
	\label{fig:model}
	\vspace{-5mm}
\end{figure}
%
The SCN extends each weight matrix of the conventional RNN to be an ensemble of a set of tag-dependent weight matrices, subjective to the probabilities that the tags are present in the image. Specifically, the SCN-RNN computes the hidden states as follows 
%
%
\begin{align} \label{eq:factored_rnn}
\hv_t = \sigma (\Wmat(\sv) \xv_{t-1} + \Umat(\sv) \hv_{t-1} + \zv ) \,, 
\end{align}
where $\zv = \mathbb{1}(t=1)\cdot \Cmat \vv$, and $\Wmat(\sv)$ and $\Umat(\sv)$ are ensembles of tag-dependent weight matrices, subjective to the probabilities that the tags are present in the image, according to the semantic-concept vector $\sv$.


%
Given $\sv \in \R^K$, we define two weight tensors $\Wmat_{\Tcal} \in \R^{n_h \times n_x \times K}$ and $\Umat_{\Tcal} \in \R^{n_h \times n_h \times K}$, where $n_h$ is the number of hidden units and $n_x$ is the dimension of word embedding. $\Wmat(\sv) \in \R^{n_h \times n_x}$ and $\Umat(\sv) \in \R^{n_h \times n_h}$ can be specified as 
\begin{align}\label{eq:tensor}
\Wmat(\sv) = \sum_{k=1}^K s_k \Wmat_{\Tcal}[k], \, 
\Umat(\sv) = \sum_{k=1}^K s_k \Umat_{\Tcal}[k] \,,
\end{align} 
where $s_k$ is the $k$-th element in $\sv$; $\Wmat_{\Tcal}[k]$ and $\Umat_{\Tcal}[k]$ denote the $k$-th 2D ``slice'' of $\Wmat_{\Tcal}$ and $\Umat_{\Tcal}$, respectively. The probability of the $k$-th semantic concept, $s_k$, is associated with a pair of RNN weight matrices $\Wmat_{\Tcal}[k]$ and $\Umat_{\Tcal}[k]$, implicitly specifying $K$ RNNs in total. Consequently, training such a model as defined in (\ref{eq:factored_rnn}) and (\ref{eq:tensor}) can be interpreted as {\em jointly} training an ensemble of $K$ RNNs. 


Though appealing, the number of parameters is proportional to $K$, which is prohibitive for large $K$ (\emph{e.g.}, $K=1000$ for COCO). In order to remedy this problem, we adopt ideas from~\cite{memisevic2007unsupervised} to factorize $\Wmat(\sv)$ and $\Umat(\sv)$ defined in (\ref{eq:tensor}) as
\begin{align}
\Wmat(\sv) &= \Wmat_a \cdot \mbox{diag}(\Wmat_b \sv) \cdot \Wmat_c \,,  \label{eq:factor_1} \\
\Umat(\sv) &= \Umat_a \cdot \mbox{diag}(\Umat_b \sv) \cdot \Umat_c \,, \label{eq:factor_2}
\end{align}
where $\Wmat_a \in \R^{n_h\times n_f}$, $\Wmat_b \in \R^{n_f \times K}$ and $\Wmat_c \in \R^{n_f \times n_x}$. Similiarly, $\Umat_a \in \R^{n_h\times n_f}$, $\Umat_b \in \R^{n_f \times K}$ and $\Umat_c \in \R^{n_f \times n_h}$. $n_f$ is the number of factors. Substituting (\ref{eq:factor_1}) and (\ref{eq:factor_2}) into (\ref{eq:factored_rnn}), we obtain our SCN with an RNN as 
\begin{align}
\tilde{\xv}_{t-1} &= \Wmat_b\sv \odot \Wmat_c\xv_{t-1} \,, \label{eq:f} \\
\tilde{\hv}_{t-1} &= \Umat_b\sv \odot \Umat_c\hv_{t-1} \,, \label{eq:g} \\
\zv &= \mathbb{1}(t=1) \cdot\Cmat \vv\,, \label{eq:b} \\
\hv_t &= \sigma (  \Wmat_a \tilde{\xv}_{t-1} + \Umat_a \tilde{\hv}_{t-1} + \zv )\,. \label{eq:h}
\end{align}
where $\odot$ denotes the element-wise multiply (Hadamard) operator.

$\Wmat_a$ and $\Wmat_c$ are shared among all the captions, effectively capturing common linguistic patterns; while the diagonal term, $\mbox{diag}(\Wmat_b\sv)$, accounts for semantic aspects of the image under test, captured by $\sv$. The same analysis also holds true for $\Umat_{a,b,c}$. In this factorized model, the RNN weight matrices that correspond to each semantic concept share ``structure.'' This factorized model (termed SCN-RNN) is illustrated in Figure~\ref{fig:model}{(b).

To provide further motivation for and insight into the decompositions in (\ref {eq:factor_1}) and (\ref {eq:factor_2}), let $\wv_{bk}$ represent the $k$th column of $\Wmat_b$, then
\begin{align}
\Wmat(\sv)= \textstyle{\sum_{k=1}^K} s_k [\Wmat_a\cdot \mbox{diag}(\wv_{bk})\cdot \Wmat_c] \,.
\end{align}
A similar decomposition is manifested for $\Umat(\sv)$. 
The matrix $\Wmat_a\cdot \mbox{diag}(\wv_{bk})\cdot \Wmat_c$ may be interpreted as the $k$-th ``slice'' of a weight tensor, with each slice corresponding to one of the $K$ semantic concepts ($K$ total tensor ``slices,'' each of size $n_h\times n_x$). Hence, via the decomposition in (\ref {eq:factor_1}) and (\ref {eq:factor_2}), we effectively learn an ensemble of $K$ sets of RNN parameters, one for each semantic concept. This is efficiently done by sharing $\Wmat_a$ and $\Wmat_c$ when composing each member of the ensemble. The weight with which the $k$-th slice of this tensor contributes to the RNN parameters for a given image is dependent on the respective probability $s_k$ with which the $k$-th semantic concept is inferred to be associated with image $\Imat$. 

The number of parameters in the basic RNN model (see Figure~\ref{fig:model}(a)) is $n_h \cdot (n_x + n_h)$, while the number of parameters in the SCN-RNN model (see Figure~\ref{fig:model}(b)) is $n_f  \cdot  (n_x+2K+3n_h)$. In experiments, we set $n_f=n_h$. Therefore, the additional number of parameters is $2\cdot n_h\cdot (n_h+K)$. This increased model complexity also indicates increased training/testing time.

\begin{table*}
	\begin{center}
		\begin{tabular}{|l|cccccc|ccccc|}
			\hline
			\multirow{2}{*}{Methods}& \multicolumn{6}{c|}{COCO} & \multicolumn{5}{c|}{Flickr30k}\\
			\cline{2-12}
			& B-1 & B-2 & B-3 & B-4 & M & C & B-1 & B-2 & B-3 & B-4 & M\\
			\hline
			NIC~\cite{vinyals2015show} & 0.666 & 0.451 & 0.304 & 0.203 & $-$ & $-$ & 0.663 & 0.423 & 0.277 & 0.183 & $-$\\
			m-RNN~\cite{mao2014deep} & 0.67 & 0.49 & 0.35 & 0.25 & $-$ & $-$ & 0.60 & 0.41 & 0.28 & 0.19 & $-$\\				
			Hard-Attention~\cite{xu2015show}  & 0.718 &  0.504 & 0.357 & 0.250 & 0.230 & $-$ & 0.669 & 0.439 & 0.296 & 0.199 & 0.185 \\
			ATT~\cite{you2016image} & 0.709 & 0.537 & 0.402 & 0.304 & 0.243 & $-$ & 0.647 & 0.460 & 0.324 & 0.230 & 0.189 \\
			Att-CNN+LSTM~\cite{wu2015value} & 0.74 & 0.56 & 0.42 & 0.31 & 0.26 & 0.94 & 0.73 & 0.55 & 0.40 & 0.28 & $-$ \\
			\hline			
			LSTM-R & 0.698 & 0.525 & 0.390 & 0.292  & 0.238 & 0.889 & 0.657 & 0.437 & 0.296 & 0.201 & 0.186 \\
			LSTM-T & 0.716 & 0.546 & 0.411 & 0.312 & 0.250 & 0.952 & 0.691 & 0.483 & 0.336 & 0.232 & 0.202 \\
			LSTM-RT & 0.724 & 0.555 & 0.419 & 0.316 & 0.252 & 0.970 & 0.706 & 0.486 & 0.339 & 0.235 & 0.204 \\
			LSTM-RT$_2$ & 0.730 & 0.568 & 0.430 & 0.322 & 0.249 & 0.977 & 0.724 & 0.523 & 0.370 & 0.257 & 0.210\\
			\hline			
			SCN-LSTM & 0.728 & 0.566 & 0.433 & 0.330 & 0.257 & 1.012 & 0.735 & 0.530 & 0.377& 0.265& 0.218 \\
			SCN-LSTM Ensemble of 5 & \textbf{0.741} & \textbf{0.578} & \textbf{0.444} & \textbf{0.341} & \textbf{0.261} & \textbf{1.041} & \textbf{0.747} & \textbf{0.552} & \textbf{0.403} & \textbf{0.288} & \textbf{0.223}\\
			\hline
		\end{tabular}
	\end{center}
	\vspace{-6mm}
	\caption{\small{Performance of the proposed model (SCN-LSTM) and other state-of-the-art methods on the COCO and Flickr30k datasets, where B-$N$, M and C are short for BLEU-$N$, METEOR and CIDEr-D scores, respectively.}}\label{tab:coco_flickr30k}
	\vspace{-4mm}
\end{table*}

\subsection{SCN-LSTM} \label{sec:img_cap_f_lstm}

RNNs with LSTM units~\cite{hochreiter1997long} have emerged as a popular architecture, due to their representational power and effectiveness at capturing long-term dependencies.  
We generalize the SCN-RNN model by using LSTM units. Specifically, we define $\hv_t = \Hcal(\xv_{t-1},\hv_{t-1}, \vv,\sv)$ as
\begin{align}
\iv_t &= \sigma ( \Wmat_{ia} \tilde{\xv}_{i,t-1} + \Umat_{ia}  \tilde{\hv}_{i,t-1} + \zv ) \,, \label{eq:i}\\
\fv_t &= \sigma ( \Wmat_{fa} \tilde{\xv}_{f,t-1} + \Umat_{fa}  \tilde{\hv}_{f,t-1} + \zv ) \,, \\
\ov_t &= \sigma ( \Wmat_{oa} \tilde{\xv}_{o,t-1} + \Umat_{oa}  \tilde{\hv}_{o,t-1} + \zv ) \,, \label{eq:o}\\
\tilde{\cv}_t &= \sigma ( \Wmat_{ca} \tilde{\xv}_{c,t-1} + \Umat_{ca}  \tilde{\hv}_{c,t-1} + \zv ) \,, \\
\cv_t &= \iv_t \odot \tilde{\cv}_t + \fv_t \odot \cv_{t-1} \,, \\
\hv_t &= \ov_t \odot \tanh (\cv_t)\,,
\end{align}
where $\zv = \mathbb{1}(t=1) \cdot \Cmat \vv$. For $\star = i,f,o,c$, we define
\begin{align}
\tilde{\xv}_{\star,t-1} &= \Wmat_{\star b} \sv \odot \Wmat_{\star c} \xv_{t-1} \,, \label{eq:mul_1}\\
\tilde{\hv}_{\star,t-1} &= \Umat_{\star b} \sv \odot \Umat_{\star c} \hv_{t-1} \,. \label{eq:mul_2}
\end{align}
Since we implement the SCN with LSTM units, we name this model SCN-LSTM. In experiments, since LSTM is more powerful than classifical RNN, we only report results using SCN-LSTM.

In summary, distinct from previous image-captioning methods, our model has a unique way to utilize and combine the visual feature $\vv$ and semantic-concept vector $\sv$ extracted from an image $\Imat$. $\vv$ is fed into the LSTM to initialize the first step, which is expected to provide the LSTM an overview of the image content. While the LSTM state is initialized with the overall visual context $\vv$, an ensemble of $K$ sets of LSTM parameters is utilized when decoding, weighted by the semantic-concept vector $\sv$, to generate the caption.

\paragraph{Model learning} Given the image $\Imat$ and associated caption $\Xmat$, the objective function is the sum of the log-likelihood of the caption conditioned on the image representation:
\begin{align}
\log p(\Xmat|\Imat) = \textstyle{\sum_{t=1}^T} p(\xv_t|\xv_0,\ldots,\xv_{t-1},\vv,\sv) \,.
\end{align}
The above objective corresponds to a single image-caption pair. In training, we average over all training pairs.

\subsection{Extension to video captioning} \label{sec:video_cap}
The above framework can be readily extended to the task of video captioning~\cite{donahue2015long,venugopalan2015sequence,venugopalan2014translating,yu2016video,ballas2015delving,xu2016msr}. In order to effectively represent the spatiotemporal visual content of a video, 
we use a two-dimensional (2D) {\em and} a three-dimensional (3D) CNN to extract visual features of video frames/clips.
We then perform a mean pooling process \cite{venugopalan2014translating} over all 2D CNN features and 3D CNN features, to generate two feature vectors (one from 2D CNN features and the other from 3D CNN features). The representation of each video, $\vv$, is produced by concatenating these two features. 
Similarly, we also obtain the semantic-concept vector $\sv$ by running the semantic-concept detector based on the video representation $\vv$. After $\vv$ and $\sv$ are obtained, we employ the same model proposed above directly for video-caption generation, as described in Figure~\ref{fig:model}{(b). 
\section{Experiments}
\subsection{Datasets}
We present results on three benchmark datasets: COCO~\cite{lin2014microsoft}, Flickr30k~\cite{young2014image} and Youtube2Text~\cite{chen2011collecting}. COCO and Flickr30k are for image captioning, containing 123287 and 31783 images, respectively. Each image is annotated with at least 5 captions. We use the same pre-defined splits as \cite{karpathy2015deep} for all the datasets: on Flickr30k, 1000 images for validation, 1000 for test, and the rest for training; and for COCO, 5000 images are used for both validation and testing. We further tested our model on the official COCO test set consisting of 40775 images (human-generated captions for this split are not publicly available), and evaluated our model on the COCO evaluation server. We also follow the publicly available code~\cite{karpathy2015deep} to preprocess the captions, yielding vocabulary sizes of 8791 and 7414 for COCO and Flickr30k, respectively.

Youtube2Text is used for video captioning, which contains 1970 Youtube clips, and each video is annotated with around 40 sentences. We use the same splits as provided in~\cite{venugopalan2014translating}, with 1200 videos for training, 100 videos for validation, and 670 videos for testing. We convert all captions to lower case and remove the punctuation, yielding vocabulary size of 12594 for Youtube2Text.

\begin{table*}
	\small
	\begin{center}
		\begin{tabular}{|l|c|c|c|c|c|c|c|c|c|c|c|c|c|c|}
			\hline
			\multirow{2}{*}{Model}& \multicolumn{2}{c|}{BLEU-1} & \multicolumn{2}{c|}{BLEU-2} & \multicolumn{2}{c|}{BLEU-3} & \multicolumn{2}{c|}{BLEU-4} & \multicolumn{2}{c|}{METEOR} & \multicolumn{2}{c|}{ROUGE-L} & \multicolumn{2}{c|}{CIDEr-D}\\
			\cline{2-15}
			& c5 & c40 & c5 & c40 & c5 & c40 & c5 & c40 & c5 & c40 & c5 & c40 & c5 & c40\\
			\hline
			SCN-LSTM & \textbf{0.740} & \textbf{0.917} & \textbf{0.575} & \textbf{0.839} & \textbf{0.436} & \textbf{0.739} & \textbf{0.331} & \textbf{0.631} & \textbf{0.257} & \textbf{0.348} & \textbf{0.543} & \textbf{0.696} & \textbf{1.003} & \textbf{1.013}\\
			\hline
			ATT & 0.731 & 0.900 & 0.565 & 0.815 & 0.424 & 0.709 & 0.316 & 0.599 & 0.250 & 0.335 & 0.535 & 0.682 & 0.943 & 0.958\\
			\hline
			OV & 0.713 & 0.895 & 0.542 & 0.802 & 0.407 & 0.694 & 0.309 & 0.587 & 0.254 & 0.346 & 0.530 & 0.682 & 0.943 & 0.946\\
			\hline
			MSR Cap& 0.715 & 0.907 & 0.543 & 0.819 & 0.407 & 0.710 & 0.308 & 0.601 & 0.248 & 0.339 & 0.526 & 0.680 & 0.931 & 0.937\\
			\hline
		\end{tabular}
	\end{center}
	\vspace{-6mm}
	\caption{\small{Comparison to published state-of-the-art image captioning models on the blind test set as reported by the COCO test server. SCN-LSTM is our model. ATT refers to ATT VC~\cite{you2016image}, OV refers to OriolVinyals~\cite{vinyals2015show}, and MSR Cap refers to MSR Captivator~\cite{devlin2015language}.}}
	\label{tab:coco_test_server}
	\vspace{-4mm}
\end{table*}

\subsection{Training procedure}
For image representation, we take the output of the 2048-way $pool5$ layer from ResNet-152~\cite{he2015deep}, pretrained on the ImageNet dataset~\cite{russakovsky2015imagenet}. For video representation, in addition to using the 2D ResNet-152 to extract features on each video frame, we also utilize a 3D CNN (C3D)~\cite{tran2014learning} to extract features on each video. The C3D is pretrained on Sports-1M video dataset~\cite{karpathy2014large}, and we take the output of the 4096-way $fc7$ layer from C3D as the video representation. We consider the RGB frames of videos as input, with 2 frames per second. Each video frame is resized as $112\times 112$ and $224\times 224$ for the C3D and ResNet-152 feature extractor, respectively. The C3D feature extractor is applied on video clips of length 16 frames (as in~\cite{karpathy2014large}) with an overlap of 8 frames.

We use the procedure described in Section~\ref{sec:visual_concept_detection} for semantic concept detection. The semantic-concept vocabulary size is determined to reflect the complexity of the dataset, which is set to 1000, 200 and 300 for COCO, Flickr30k and Youtube2Text, respectively. Since Youtube2Text is a relatively small dataset, we found that it is very difficult to train a reliable semantic-concept detector using the Youtube2Text dataset alone, due to its limited amount of data. In experiments, we utilize additional training data from COCO. 

For model training, all the parameters in the SCN-LSTM are initialized from a uniform distribution in [-0.01,0.01].  All bias terms are initialized to zero.  Word embedding vectors are initialized with the publicly available \emph{word2vec} vectors~\cite{mikolov2013distributed}. The embedding vectors of words not present in the pretrained set are initialzied randomly. The number of hidden units and the number of factors in SCN-LSTM are both set to 512 and we use mini-batches of size 64. The maximum number of epochs we run for all the three datasets is 20.
Gradients are clipped if the norm of the parameter vector exceeds 5~\cite{sutskever2014sequence}. We do not perform any dataset-specific tuning and regularization other than dropout~\cite{zaremba2014recurrent} and early stopping on validation sets. 
The Adam algorithm~\cite{kingma2014adam} with learning rate $2\times10^{-4}$ is utilized for optimization. All experiments are implemented in Theano~\cite{2016arXiv160502688short}\footnote{Code is publicly available at \url{https://github.com/zhegan27/Semantic_Compositional_Nets}.}.

In testing, we use beam search for caption generation, which selects the top-$k$ best sentences at each time step and considers them as the candidates to generate new top-$k$ best sentences at the next time step. We set the beam size to $k=5$ in experiments.

\subsection{Evaluation}

\begin{table}
	\small
	\begin{center}
		\begin{tabular}{|l|ccc|}
			\hline
			Model  & B-4 & M & C \\
			\hline
			S2VT~\cite{venugopalan2015sequence} & $-$ & 0.292 & $-$ \\
			LSTM-E~\cite{pan2016joint}  & 0.453 & 0.310 & $-$ \\
			GRU-RCN~\cite{ballas2015delving}  & 0.479 & 0.311 & 0.678 \\
			h-RNN~\cite{yu2016video}  & 0.499 & 0.326 & 0.658 \\
			\hline
			LSTM-R  & 0.448 & 0.310 & 0.640 \\
			LSTM-C  & 0.445 & 0.309 & 0.644 \\
			LSTM-CR  & 0.469 & 0.317 & 0.688 \\
			LSTM-T  & 0.473 & 0.324 & 0.699 \\
			LSTM-CRT  & 0.475 & 0.316 & 0.647 \\
			LSTM-CRT$_2$  & 0.469 & 0.326 & 0.706 \\
			\hline
			SCN-LSTM  & 0.502 & 0.334 & 0.770 \\
			SCN-LSTM Ensemble of 5  & \textbf{0.511} & \textbf{0.335} & \textbf{0.777} \\
			\hline
		\end{tabular}
	\end{center}
	\vspace{-6mm}
	\caption{\small{Results on BLEU-4 (B-4), METEOR (M) and CIDEr-D (C) metrices compared to other state-of-the-art results and baselines on Youtube2Text.}}
	\label{tab:youtube}
	\vspace{-5mm}
\end{table}

The widely used BLEU \cite{papineni2002bleu}, METEOR~\cite{banerjee2005meteor}, ROUGE-L~\cite{lin2004rouge}, and CIDEr-D~\cite{vedantam2015cider} metrics are reported in our quantitative evaluation of the performance of the proposed model and baselines in the literature. All the metrics are computed by using the code released by the COCO evaluation server~\cite{chen2015microsoft}. For COCO and Flickr30k datasets, besides comparing to results reported in previous work, we also re-implemented strong baselines for comparison. The results of image captioning are presented in Table~\ref{tab:coco_flickr30k}. The models we implemented are as follows.

\begin{enumerate}
	\item
	{\em LSTM-R / LSTM-T / LSTM-RT}: {\em R, T, RT} denotes using different features. Specifically, {\em R} denotes ResNet visual feature vector, {\em T} denotes Tags (\emph{i.e.}, the semantic-concept vector),  and {\em RT} denotes the concatenation of {\em R} and {\em T}. The features are fed into a standard LSTM decoder only at the initial time step. 
	In particular, {\em LSTM-T} is the model proposed in~\cite{wu2015value}.
	\vspace{-1.1mm}
	\item
	{\em LSTM-RT$_2$}:  The ResNet feature vector is sent to a standard LSTM decoder at the first time step, while the tag vector is sent to the LSTM decoder at every time step in addition to the input word. This model is  similar to~\cite{you2016image} without using semantic attention.
	This is the model closest to ours, which provides a direct comparison to our proposed model. 
	\vspace{-1.1mm}
	\item
	{\em SCN-LSTM}: This is the model presented in Section~\ref{sec:img_cap_f_lstm}. 
\end{enumerate}

For video captioning experiments, we use the same notation. For example, {\em LSTM-C} means we leverage the C3D feature for caption generation. 

\begin{figure*}[th!]
	\centering
	\includegraphics[width=1.00\textwidth]{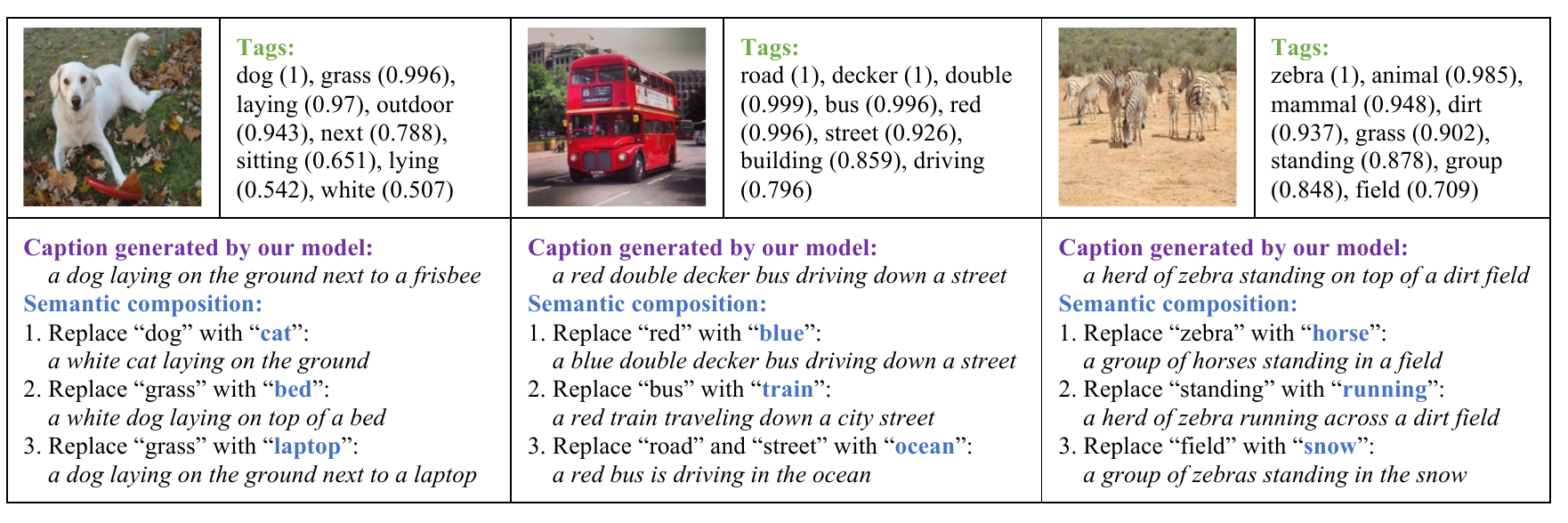}
	\vspace{-7mm}
	\caption{\small{Illustration of semantic composition. Our model can adjust the caption smoothly as the semantic concepts are modified. }}
	\vspace{-3mm}
	\label{fig:example_semantic_composition}
\end{figure*}

\begin{figure*}[th!]
	\centering
	\includegraphics[width=1.00\textwidth]{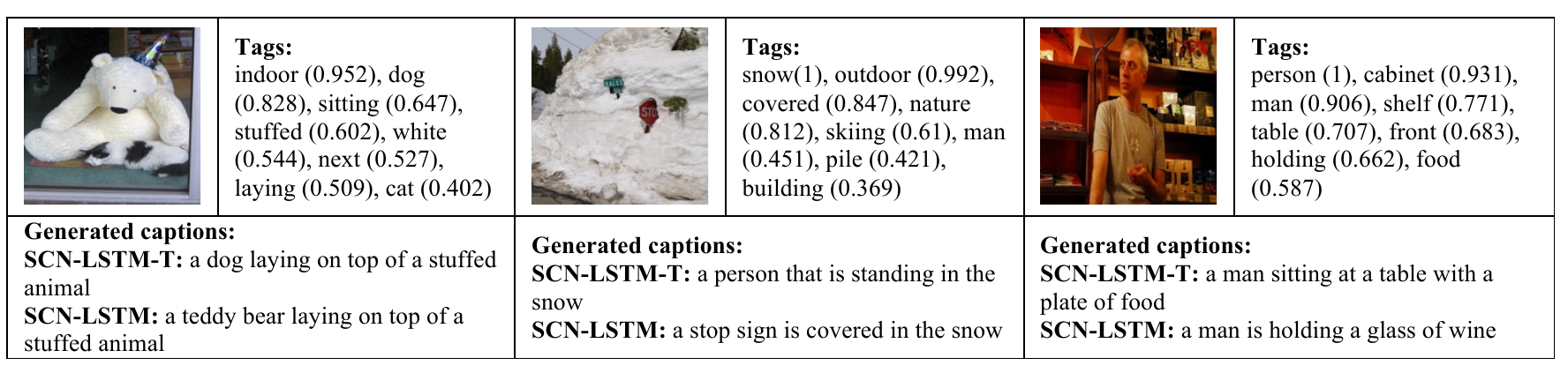}
	\vspace{-7mm}
	\caption{\small{Detected tags and sentence generation results on COCO. The output captions are generated by: 1) SCN-LSTM, and 2) SCN-LSTM-T, a SCN-LSTM model without the visual feature inputs, \ie, with only tag inputs.}}
	\vspace{-4mm}
	\label{fig:example_for_rebuttal}
\end{figure*}

\subsection{Quantitative results}
\paragraph{Performance on COCO and Flickr30k} We first present results on the task of image captioning, summarized in Table~\ref{tab:coco_flickr30k}. The use of tags ({\em LSTM-T}) provides better performance than leveraging visual features alone ({\em LSTM-R}). Combining both tags and visual features further enhances performance, as expected.  Compared with only feeding the tags into the LSTM at the initial time step ({\em LSTM-RT}), {\em LSTM-RT$_2$} yields better results, since it takes as input the tag feature at each time step. Further, the direct comparison between {\em LSTM-RT$_2$} and {\em SCN-LSTM} demonstrates the advantage of our proposed model, indicating that our approach is a better method to fuse semantic concepts into the LSTM. 

We also report results averaging an ensemble of 5 identical {\em SCN-LSTM} models trained with different initializations, which is a common strategy adopted widely~\cite{you2016image} 
(note that now we employ ensembles in two ways: an ensemble of LSTM parameters linked to tags, and an overaching ensemble atop the entire model). 
We obtain state-of-the-art results on both COCO and Flickr30k datasets. Remarkably, we improve the state-of-the-art BLEU-4 score by 3.1 points on COCO. 

\vspace{-4mm}
\paragraph{Performance on COCO test server} We also evaluate the proposed \emph{SCN-LSTM} model by uploading results to the online COCO test server. Table~\ref{tab:coco_test_server} shows the comparison to the published state-of-the-art image captioning models on the blind test set as reported by the COCO test server. We include the models that have been published and perform at top-3 in the table. Compared to these methods, our proposed {\em SCN-LSTM} model achieves the best performance across all the evaluation metrics on both c5 and c40 testing sets.\footnote{Please check \url{https://competitions.codalab.org/competitions/3221\#results} for the most recent results.} 

\vspace{-4mm}
\paragraph{Performance on Youtube2Text} Results on video captioning are presented in Table~\ref{tab:youtube}. The SCN-LSTM achieves significantly better results over all competing methods in all metrics, especially in CIDEr-D. For self-comparison, it is also worth noting that our model improves over {\em LSTM-CRT$_2$} by a substantial margin. Again, using an overaching ensemble further enhances performance.

\subsection{Qualitative analysis}
\begin{figure*}[th!]
	\centering
	\includegraphics[width=1.00\textwidth]{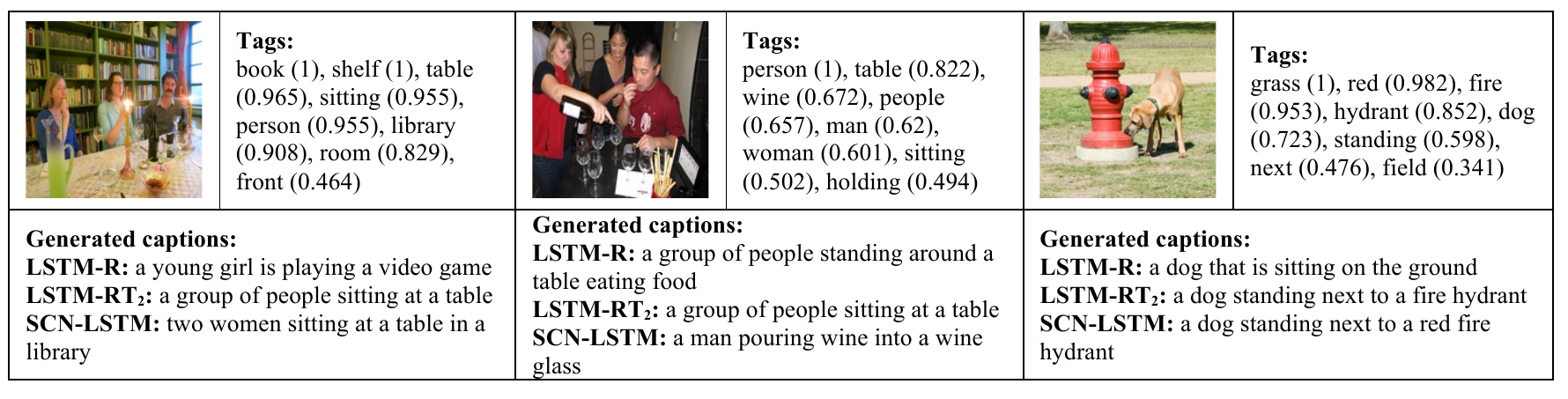}
	\vspace{-7mm}
	\caption{\small{Detected tags and sentences generation results on COCO. The output captions are generated by: 1) LSTM-R, 2) LSTM-RT$_2$, and 3) our SCN-LSTM.}}
	\label{fig:example_img_cap}
	\vspace{-3.5mm}
\end{figure*}

\begin{figure*}[th!]
	\centering
	\includegraphics[width=1.00\textwidth]{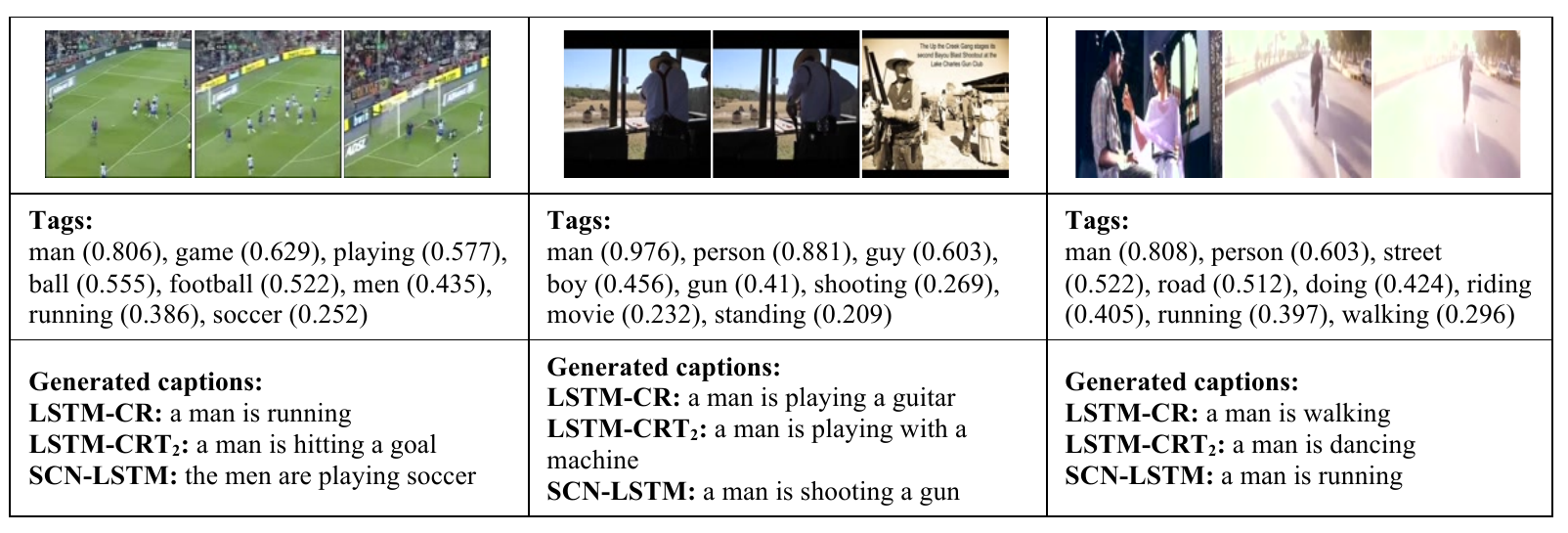}
	\vspace{-7mm}
	\caption{\small{Detected tags and sentence generation results on Youtube2Text. The output captions are generated by: 1) LSTM-CR, 2) LSTM-CRT$_2$, and 3) our SCN-LSTM. }}
	\vspace{-4mm}
	\label{fig:example_video_cap}
\end{figure*}

Figure~\ref{fig:example_semantic_composition} shows three examples to illustrate the semantic composition on caption generation. Our model properly describes the image content by using the correctly detected tags. By manually replacing specific tags, our model can adjust the caption smoothly. For example, in the left image, by replacing the tag \emph{``grass''} with \emph{``bed''}, our model imagines \emph{``a dog laying on top of a bed''}. Our model is also able to generate novel captions that are highly unlikely to occur in real life. For instance, in the middle image, by replacing the tag \emph{``road''} and \emph{``street''} with \emph{``ocean''}, our model imagines \emph{``a bus driving in the ocean''}; in the right image, by replacing the tag \emph{``field''} with \emph{``snow''}, our model dreams \emph{``a group of zebras standing in the snow''}. 

SCN not only picks up the tags well (and imagines the corresponding scenes), but also selects the right functional words for different concepts to form syntactically correct caption. As illustrated in sentence 6 of Figure~\ref{fig:illustration}, by replacing the tag \emph{``baby''} with \emph{``girl''}, the generated captions not only changes \emph{``a baby''} to \emph{``a little girl''}, but more importantly, changes \emph{``in its mouth''} to \emph{``in her mouth''}. In addition, the SCN also infers the underlying semantic relatedness between different tags. As illustrated in sentence 4 of Figure~\ref{fig:illustration}, when only switching on the tag \emph{``mouth''}, the generated caption becomes \emph{``a man with a toothbrush''}, indicating the semantic closeness between \emph{``mouth''}, \emph{``man''} and \emph{``toothbrush''}. By further switching on \emph{``baby''}, we generate a more detailed description \emph{``a baby brushing its teeth''}.

The above analysis shows the importance of tags in generating captions. However, SCN generates captions using both semantic concepts and the global visual feature vector. The language model learns to assemble semantic concepts (weighted by their likelihood), in consideration of the global visual information, into a coherent meaningful sentence that captures the overall meaning of the image. In order to demonstrate the importance of visual feature vectors, we train another SCN-LSTM-T model, which is a SCN-LSTM model without the visual feature inputs, \ie, with only tag inputs . As shown in the first example of Figure~\ref{fig:example_for_rebuttal}, the image tagger detects ``\emph{dog}'' with high probability. Using only tag inputs, SCN-LSTM-T can only generate the wrong caption ``\emph{a dog laying on top of a stuffed animal}''. With additional visual feature inputs, our SCN-LSTM model correctly replaces ``\emph{dog}'' with ``\emph{teddy bear}'' .

We further present examples of generated captions on COCO with various other methods in Figure~\ref{fig:example_img_cap}, along with the detected tags. As can be seen, our model often generates more reasonable captions than \emph{LSTM-R}, due to the use of high-level semantic concepts. For example, in the first image, \emph{LSTM-R} outputs an irrelevant caption to the image, while the detection of \emph{``table''} and \emph{``library''} helps our model to generate more sensible caption. 
Further, although both our model and \emph{LSTM-RT$_2$} utilize detected tags for caption generation, our model often depicts the image content more comprehensively;  \emph{LSTM-RT$_2$} has a larger potential to miss important details in the image. 
For instance, in the 3rd image, the tag ``\emph{red}'' appears in the caption generated by our model, which is missed by \emph{LSTM-RT$_2$}. This observation might be due to the fact that the SCN provides a better approach to fuse tag information into the process of caption generation. 
Similiar observations can also be found in the video captioning experiments, as demonstrated in Figure~\ref{fig:example_video_cap}.

\section{Conclusion}
%
We have presented Semantic Compositional Network (SCN), a new framework to effectively compose the individual semantic meaning of tags for visual captioning. The SCN extends each weight matrix of the conventional LSTM to be a three-way matrix product, with one of these matrices dependent on the inferred tags. Consequently, the SCN can be viewed 
an ensemble of tag-dependent LSTM bases, with the contribution of each LSTM basis unit proportional to the likelihood that the tag is present in the image. 
Experiments conducted on three visual captioning datasets validate the superiority of the proposed approach.

\paragraph{Acknowledgements}
Most of this work was done when the first author was an intern at Microsoft Research.
This work was also supported in part by ARO, DARPA, DOE, NGA, ONR and NSF.

\newpage
{\small
\bibliographystyle{ieee}
\bibliography{img_cap}
}

\twocolumn[{%
	\renewcommand\twocolumn[1][]{#1}%
	\appendix
	\section{More results on image captioning}
	\begin{center}
		\centering
		\includegraphics[width=1.00\textwidth]{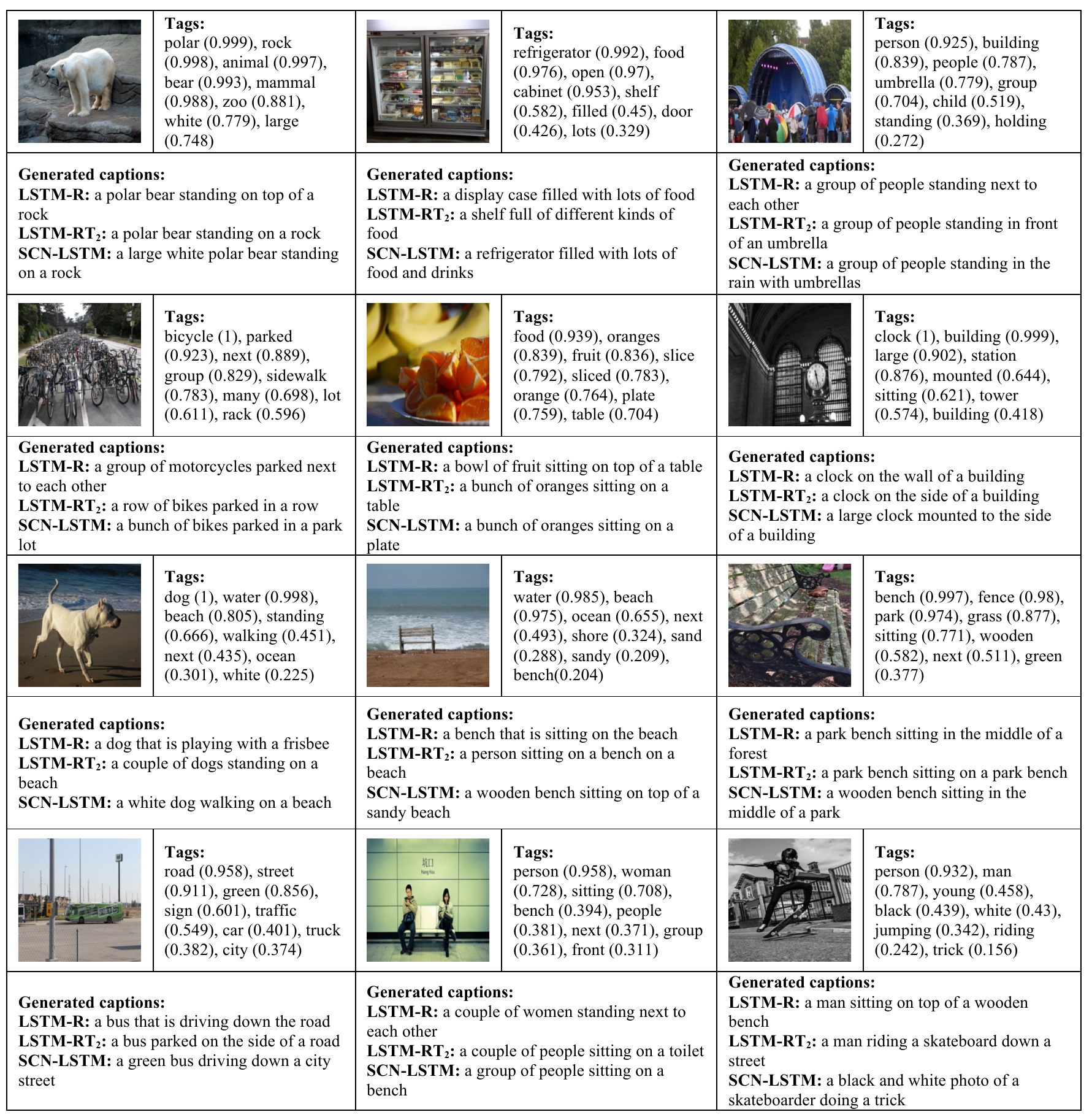}
		\captionof{figure}{\small{Detected tags and sentences generation results on COCO. The output captions are generated by: 1) LSTM-R, 2) LSTM-RT$_2$, and 3) our SCN-LSTM.}}
	\end{center}
}]

\twocolumn[{%
	\renewcommand\twocolumn[1][]{#1}%
	\section{More results on video captioning}
	\begin{center}
		\centering
		\includegraphics[width=1.00\textwidth]{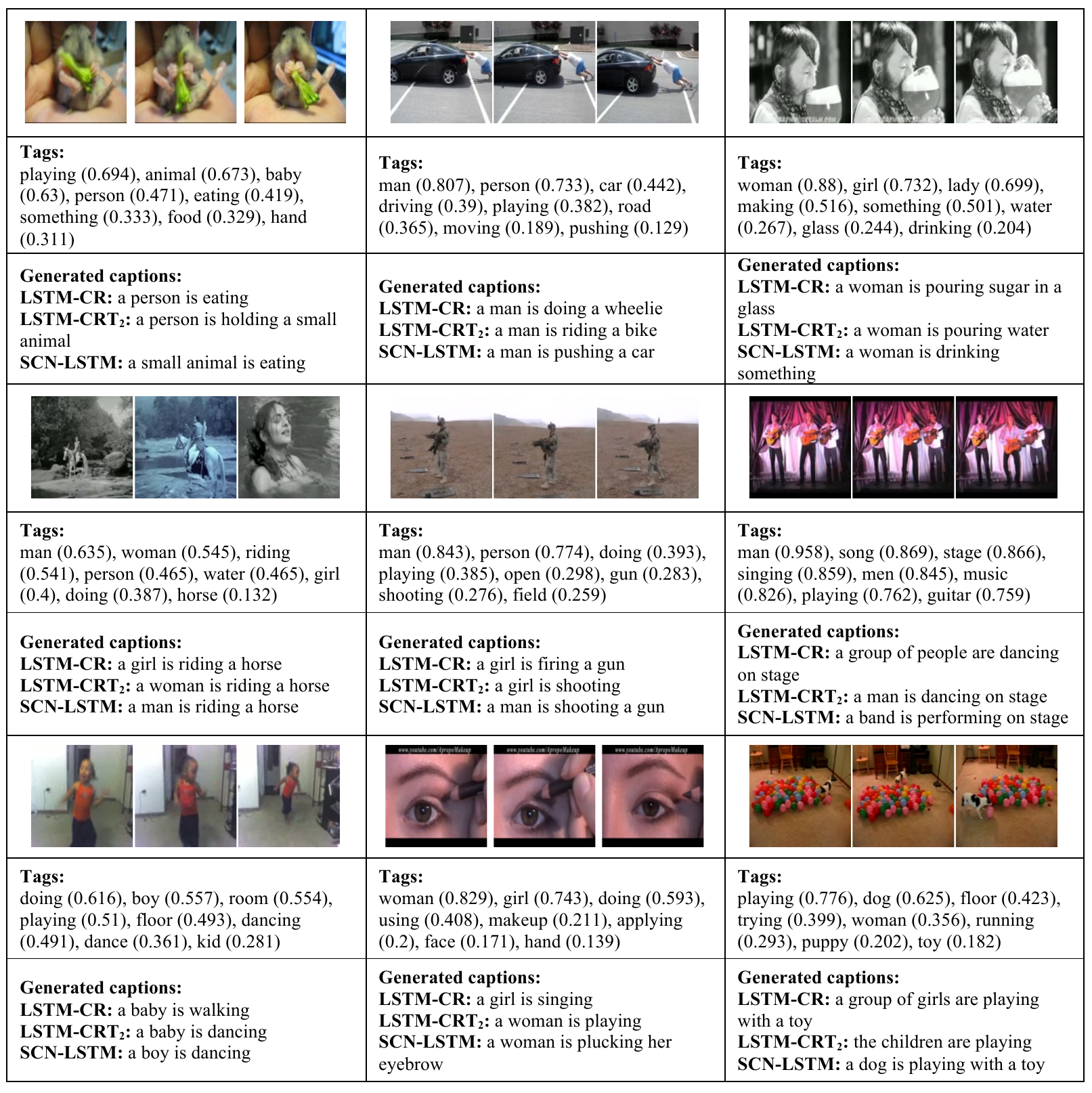}
		\captionof{figure}{\small{Detected tags and sentence generation results on Youtube2Text. The output captions are generated by: 1) LSTM-CR, 2) LSTM-CRT$_2$, and 3) our SCN-LSTM. }}
	\end{center}
}]

\twocolumn[{%
	\renewcommand\twocolumn[1][]{#1}%
	\section{More results for Figure 4}
	\begin{center}
		\centering
		\includegraphics[width=1.00\textwidth]{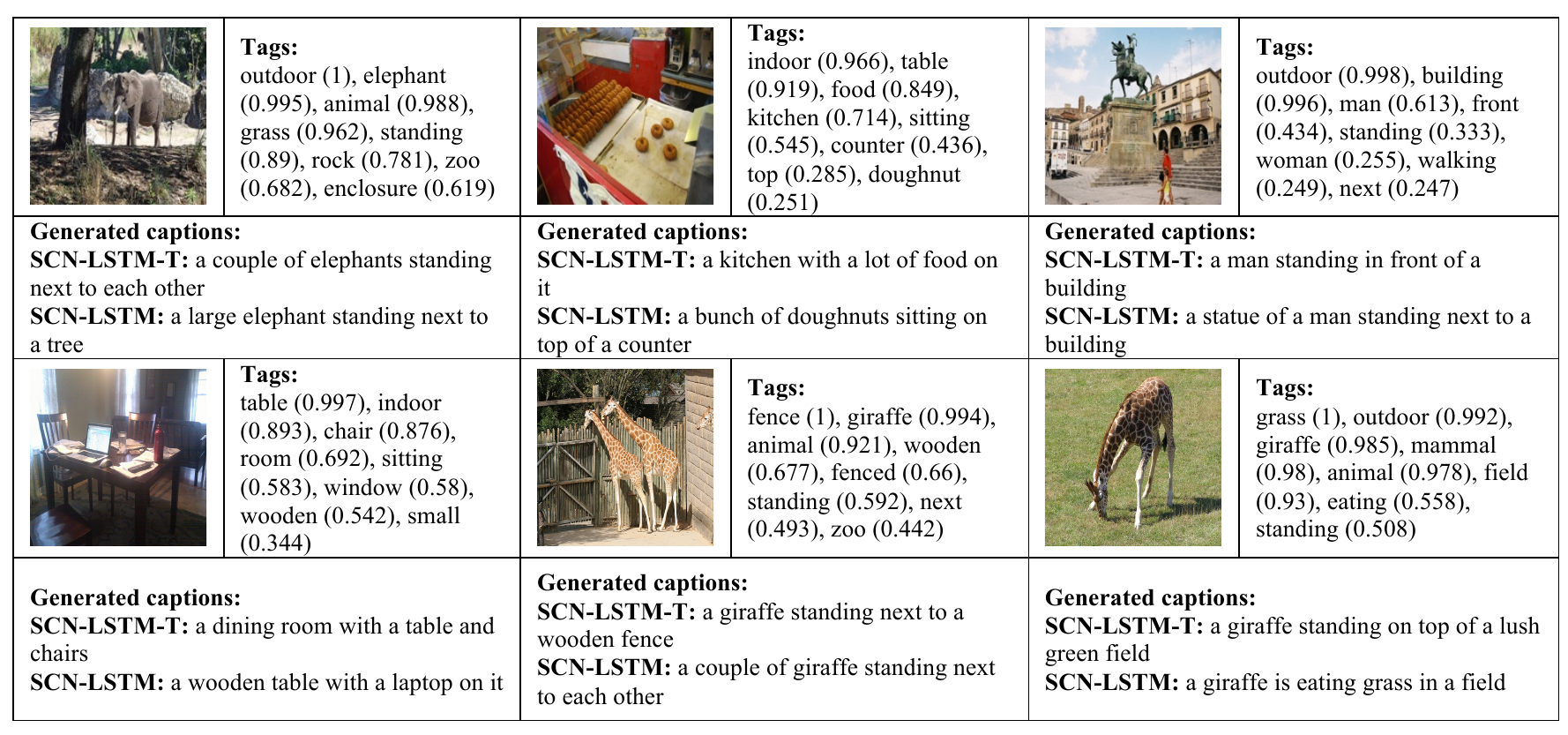}
		\captionof{figure}{\small{Detected tags and sentence generation results on COCO. The output captions are generated by: 1) SCN-LSTM, and 2) SCN-LSTM-T, a SCN-LSTM model without the visual feature inputs, \ie, with only tag inputs. }}
	\end{center}
}]

\end{document}